\pdfoutput=1
\documentclass[11pt]{article}

\usepackage[preprint]{acl}
\usepackage{graphicx}
\usepackage{subcaption} 
\newcommand{\figref}[1]{Figure ~\ref{#1}}
\usepackage{times}
\usepackage{latexsym}
\usepackage{amsmath}
\usepackage{bbding,pifont,paralist}
\usepackage{color}
\usepackage{tabularray}
\usepackage{enumerate}
\usepackage{graphicx}
\usepackage[T1]{fontenc}
\usepackage[utf8]{inputenc}

\usepackage{microtype}

\usepackage{inconsolata}

\usepackage{graphicx}

\title{A Novel Method to Metigate Demographic and Expert Bias in  ICD Coding with Causal Inference}

\author{First Author \\
  Bin Zhang \\ Tongji University, Shanghai, China \\
  \texttt{2233009@tongji.edu.cn} \\\And
  Second Author \\
  Junli Wang \\ Tongji University, Shanghai, China \\
  \texttt{junliwang@tongji.edu.cn} \\}

\begin{document}
\maketitle
\begin{abstract}
ICD(International Classification of Diseases) coding involves assigning ICD codes to patients visit based on their medical notes. Considering ICD coding as a multi-label text classification task, researchers have developed sophisticated methods. Despite progress, these models often suffer from label imbalance and may develop spurious correlations with demographic factors. Additionally, while human coders assign ICD codes, the inclusion of irrelevant information from unrelated experts introduces biases. 
To combat these issues, we propose a novel method to mitigate \textbf{D}emographic and \textbf{E}xpert biases in ICD coding through \textbf{C}ausal \textbf{I}nference (\textbf{DECI}). We provide a novel causality-based interpretation in ICD Coding that models make predictions by three distinct pathways. And based counterfactual reasoning, \textbf{DECI} mitigate demographic and expert biases. Experimental results show that DECI outperforms state-of-the-art models, offering a significant advancement in accurate and unbiased ICD coding.
\end{abstract}
\section{Introduction}

The International Classification of Diseases (ICD), managed by the World Health Organization, standardizes diagnostic and procedural data from patient encounters. These codes are essential for global health information standardization, epidemiological research, service billing, and health trend monitoring \citep{epidemiological, bill, monitor}. However, manual ICD coding is challenging due to the need for specialized training, complexity in code assignment, and is prone to errors, high costs, and time inefficiency \citep{manual}, driving interest in automated solutions.

Medical code prediction, as a multi-label text classification task, has seen the development of various machine learning approaches including rule-based systems \citep{rule}, SVMs \citep{svm}, and decision trees \citep{decision}. The advent of deep learning in NLP has led to the use of RNNs \citep{LAAT, MSMN}, CNNs \citep{caml, multirescnn}, and Transformer-based models \citep{Fusion, 2024AccurateAW, 2024KnowledgebasedDP} for learning deep representations of medical notes towards predicting ICD codes. Advanced techniques now incorporate label attention mechanisms to highlight key words within medical notes, enhancing the accuracy of code predictions \citep{2023Rare-ICD, 2024CoRelationBA}.

\begin{figure}[h]
    \centering
    \begin{subfigure}[b]{0.5\textwidth}
        \includegraphics[width=\textwidth]{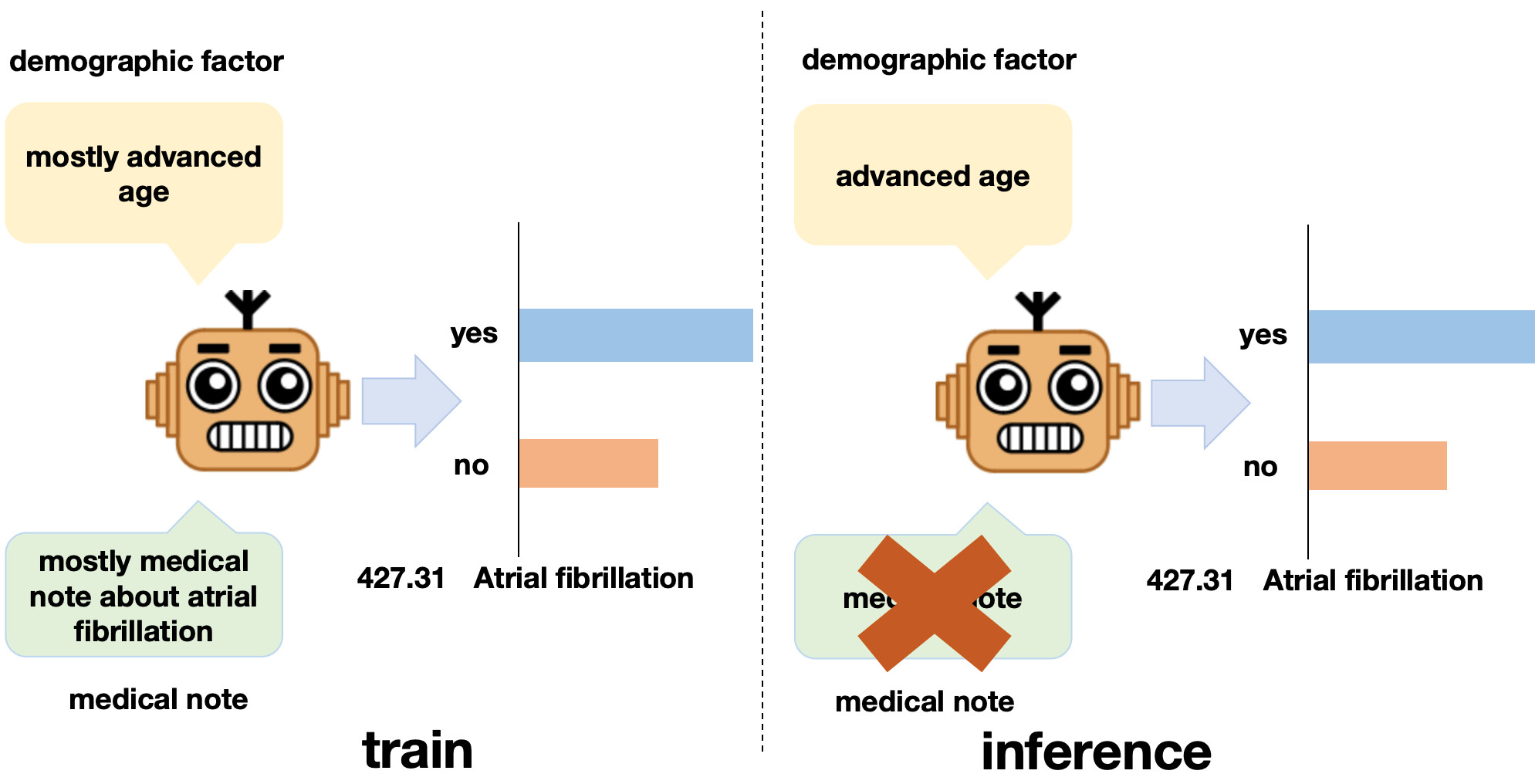}
        \caption{}
        \label{Figure 1.a}
    \end{subfigure}
    \begin{subfigure}[b]{0.5\textwidth}
        \includegraphics[width=\textwidth]{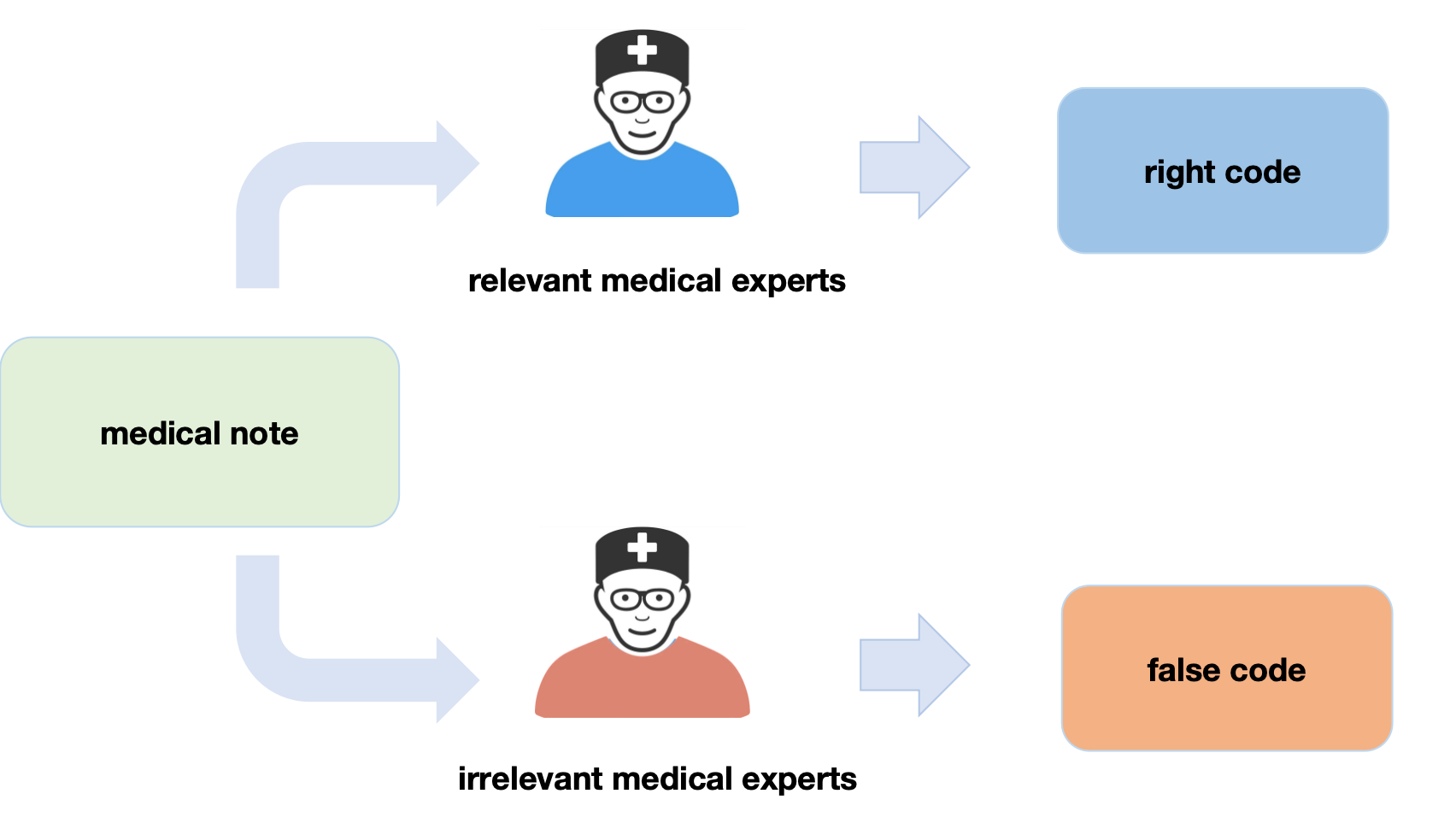}
        \caption{}
        \label{Figure 1.b}
    \end{subfigure}

    \caption{(a) An example illustrates that model overlooks the medical note and makes predictions rely on demographic factors. (b) Example of irrelevant expert in other department assign false ICD codes.}
    \label{Figure 1}
\end{figure}

Recent studies \citep{shim-etal-2022-exploratory} highlight that neural natural language processing models used for medical code prediction frequently face a label imbalance issue. These models tend to leverage these imbalances, learning shortcuts instead of focusing on the underlying task. As a result, predictions may rely heavily on demographic factors like gender and age, which can lead to spurious correlations. As shown in \figref{Figure 1} (a), in the training set, a higher co-occurrence probability was observed between advanced age and the ICD code \textit{"427.31 Atrial fibrillation"}. Consequently, during the inference phase, the model exhibited a tendency to to overlook medical note and assign the ICD code \textit{"427.31 Atrial fibrillation"} with advanced age. Given the importance of equitable service quality in healthcare settings, it is crucial to avoid learning such spurious correlations. To address this, we subtract the direct effect of demographic factors on ICD codes in inference phase. 

When human coders assign ICD codes, they typically draw upon the knowledge specific to their respective departments \citep{explore2024}. However, an expert in one department, such as hand and foot surgery, may lack the core knowledge of another, like urology. This disparity can introduce irrelevant information when assigning ICD codes. As shown in \figref{Figure 1} (b), when irrelevant medical experts are tasked with predicting codes, they tend to forecast ICD codes that are familiar to them but are not right. This inclination introduces extraneous knowledge, thereby affecting the final outcomes.. To model this expertise, we implement a Mixture of Experts (MoE) \citep{MOE} framework, representing specialists from various departments. And we subtract the direct effect of other irrelevant experts on ICD codes to mitigate the biases introduced by knowledge from unrelated departments.

Therefore, in this paper, we introduce a novel method to mitigate \textbf{D}emographic and \textbf{E}xpert biases in ICD coding through \textbf{C}ausal \textbf{I}nference (\textbf{DECI}). Specifically, we provide a novel causality-based interpretation in ICD Coding that models make predictions by three distinct pathways. To address demographic and expert biases, we incorporate counterfactual reasoning into the ICD coding process.Experimental outcomes indicate that our proposed method outperforms leading state-of-the-art models on a benchmark dataset.

The main contributions of this paper are summarized as follows:
\begin{itemize}
\item We present \textbf{DECI}, a novel method aimed at reducing demographic and expert biases in ICD coding. This is the first known attempt to specifically address demographic bias and expert bias in this context.
\item Our work provide a novel causality-based interpretation in ICD Coding that models make predictions by three distinct pathways.
\item DECI has been shown to markedly outperform current state-of-the-art methods and the in-depth analysis provides the rationality..
\end{itemize}

% \begin{itemize}
% \item We introduce a novel method, termed \textbf{D}emographic and \textbf{E}xpert \textbf{C}ausal \textbf{I}nference (\textbf{DECI}), designed to alleviate demographic and expert biases in the process of ICD coding. To our knowledge, this represents the first endeavor specifically targeting the reduction of demographic bias in ICD coding.
% \item Furthermore, this work marks an initial foray into addressing expert bias in ICD coding by implementing a MoE framework, which encapsulates the experts from a variety of medical departments. As far as we are aware, no prior studies have undertaken such an initiative.
% \item Notably, our method does not require additional text preprocessing and introduces minimal parameter overhead. Comprehensive evaluations demonstrate that it significantly improves upon existing state-of-the-art baselines.
% \end{itemize}
\section{Related Work}

The task of ICD coding has garnered considerable attention over the past several decades. In earlier efforts, \citet{rule} employed a rule-based approach to extract pertinent snippets from medical records and associate them with corresponding ICD codes. Subsequently, \citet{svm} introduced a SVM classifier that utilized bag-of-words features for this purpose. An attempt was also made by \citet{decision} to identify critical features through a decision tree-based methodology. However, these traditional methods often failed to achieve satisfactory performance, mainly due to the difficulty in extracting meaningful features from complex and noisy medical narratives.

The advent of neural networks led to an increased focus on applying advanced models such as Recurrent Neural Networks (RNNs) \citep{rnn2023,rnn2024}, Convolutional Neural Networks (CNNs) \citep{CNN2024}, and Transformer \citep{LIU2023102662,NIU2023104396,transformer2-24} architectures for ICD coding. For instance, \citep{multirescnn} implemented convolutional layers with varying kernel sizes to capture relevant information for each code from the source medical texts. Another notable work, \citep{LAAT}, focused on detecting label-specific terms within notes using Long Short-Term Memory (LSTM) networks enhanced with a tailored label attention mechanism. Meanwhile, it is pointed out in \citep{transicd} the use of Transformer-based approaches, demonstrating comparable performance to CNN-based models.

Recent studies including \citep{shim-etal-2022-exploratory}, have highlighted a common issue faced by NLP models in medical code prediction: label imbalance. These models \citep{yang2023multi, gpt2023} tend to make predictions that are heavily influenced by demographic factors, such as gender and age, which can result in spurious correlations. Considering the necessity of providing equitable healthcare services, it is imperative to avoid learning these types of correlations. To address this challenge, we propose a counterfactual reasoning intervention  to decouple the direct relationship between demographic variables and predicted labels.

Moreover, there exists a phenomenon where ICD coding can be affected by the specialized knowledge of different medical departments \citep{explore2024}. When human coders assign ICD codes \citep{icd-10}, they usually draw upon their specific departmental expert. However, this can lead to the introduction of irrelevant or biased information, as an expert in one field, like hand and foot surgery, may not possess the requisite knowledge in another, such as urology. To model this domain-specific expertise, we adopt a Mixture of Experts (MoE) framework\citep{MOE}, to represent specialists from various medical fields. Additionally, we apply counterfactual reasoning techniques to mitigate biases stemming from unrelated areas of expertise, thereby improving the accuracy and relevance of ICD code assignments.
\section{Methodology}
\subsection{Background}
In this section, we provide a succinct overview of the fundamental problem configuration for ICD coding and the foundational approach employed by the models to transform medical notes into the corresponding ICD codes.

\textbf{ICD Coding:}
The process of ICD coding\citep{multirescnn} can be characterized as a sophisticated multi-label classification challenge. Formally, let us consider a medical document \( d \) consisting of a sequence of \( N \) tokens:
\begin{equation}
    d = \{t_1, t_2, \ldots, t_{N}\},
\end{equation}
and a set of medical codes \( L \), which is defined as \( L = \{l_1, l_2, \ldots, l_{N_L}\} \), where \( N_L \) denotes the cardinality of the code set.

The objective of the multi-label classification task is to decompose it into a series of binary classification problems, with the aim of assigning a binary label \( l_i \in \{0,1\} \) to each individual ICD code in the set \( L \). A value of 1 signifies the presence of a condition or procedure that corresponds to the particular ICD code.

\textbf{Base Method:}
The base model\citep{MSMN,LAAT,2023Rare-ICD} encompasses an encoder, which is responsible for transfering the input text into token-level embeddings \begin{math}E = \text{Enc}(t_1, t_2, \ldots, t_{N})\end{math}, where \( E \) represents the encoded representation of the input tokens. Subsequently, attention mechanisms are utilized to align and refine these representations with respect to the descriptions of the potential ICD codes, yielding a label-specific document representation denoted as \( H=\text{Attention}(E) \).

Subsequent to the aforementioned alignment, the label-specific representation \( H \) is propagated through a feed-forward neural network (FFN) for the computation of scores pertaining to each ICD code, denoted as \( Z=FFN(H) \). The final step involves the prediction of the complete set of associated ICD codes, which is mathematically expressed as \( L = \sigma(Z) \), where \( \sigma \) signifies the activation function employed for this transformation.

\subsection{Proposed Framework}

\begin{figure*}[h]
    \centering
    \begin{subfigure}[b]{0.3\textwidth}
        \includegraphics[width=\textwidth]{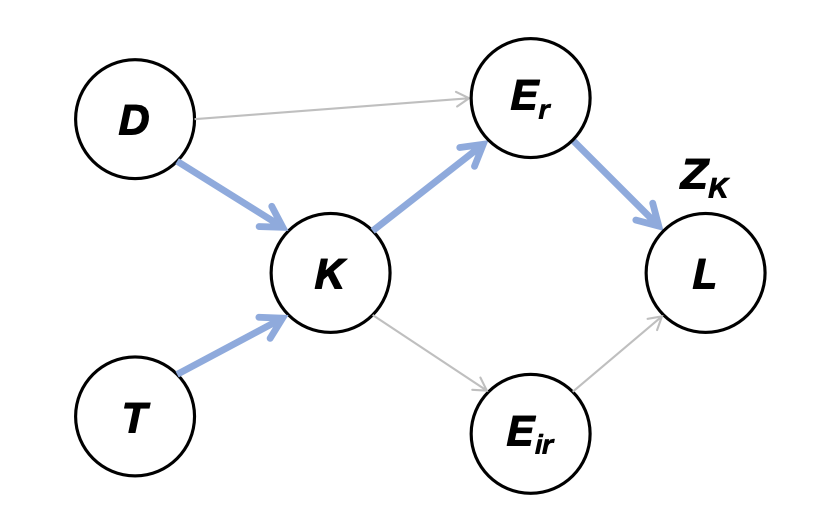}
        \label{Figure 2.a}
    \end{subfigure}
    \begin{subfigure}[b]{0.3\textwidth}
        \includegraphics[width=\textwidth]{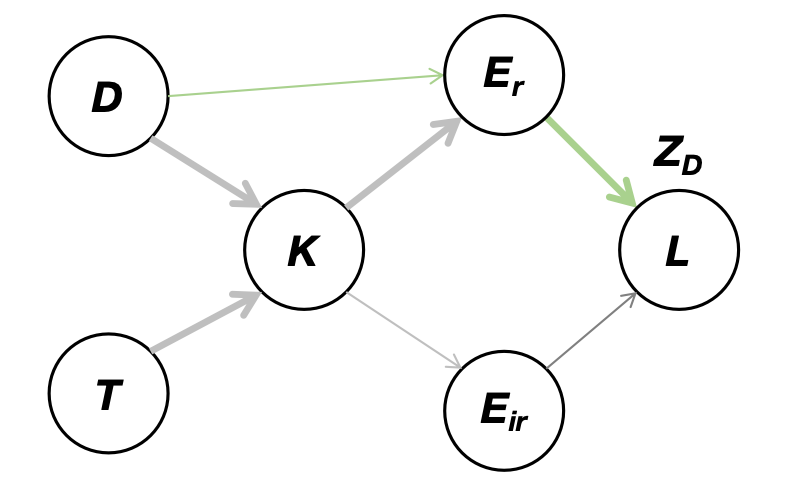}
        \caption{}
        \label{Figure 2.b}
    \end{subfigure}
    \begin{subfigure}[b]{0.3\textwidth}
    \includegraphics[width=\textwidth]{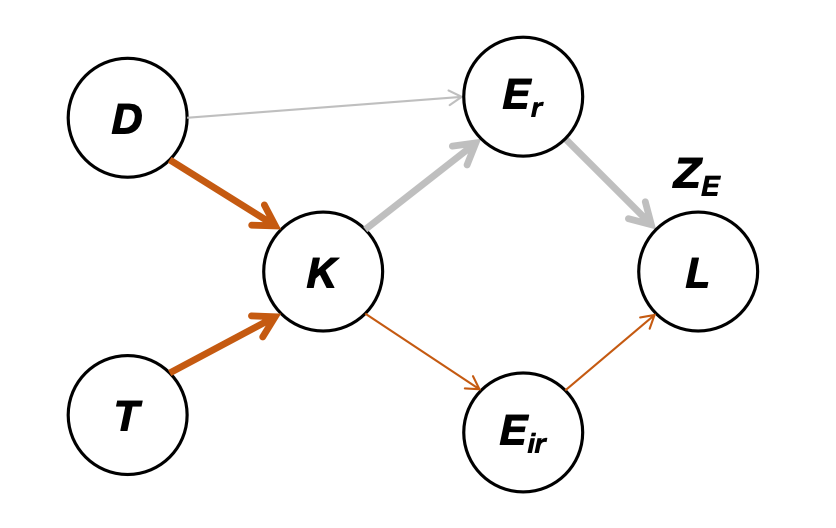}
        \label{Figure 2.c}
    \end{subfigure}

    \begin{subfigure}[h]{\textwidth}
    \centering
    \includegraphics[scale=0.4]{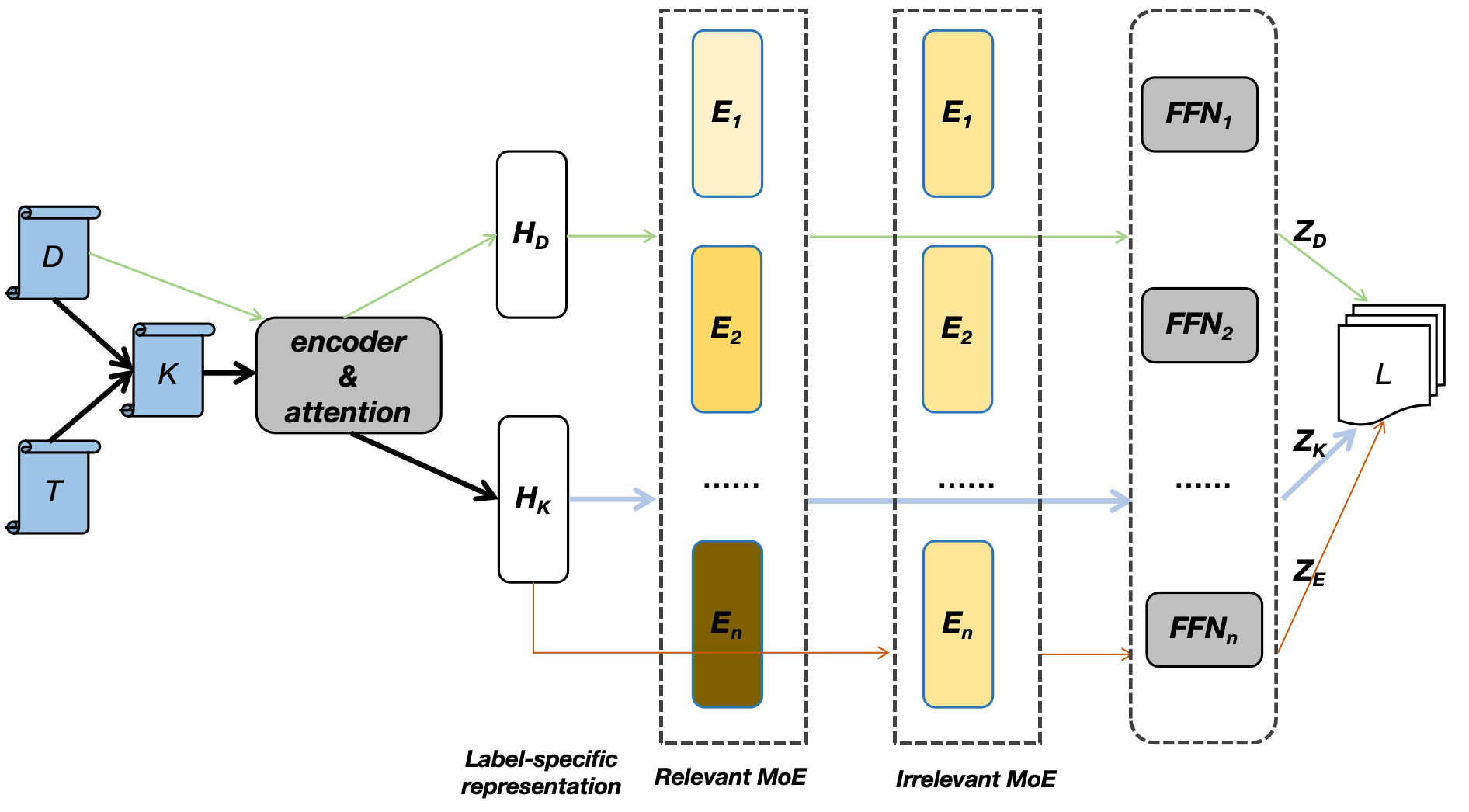} 
	\caption{}
	\label{Figure 2.b} 
    \end{subfigure}
    \caption{(a) The causal graph for ICD Coding. \( D \), \( T \), \( K \), \( L \), \( E_r \) and  \( E_\textit{ir} \) represent demographic factors, the medical note, aggregated knowledge, label,  relevant expert and  irrelevant expert respectively. \textbf{DECI} emphasizes the aggregated knowledge causal reasoning with relevant expert(blue), simultaneously eliminating effects from demographic factors (green) and irrelevant expert(red). (b) The framework of proposed \textbf{DECI} method. The degree of MoE shading assigned to individual experts denotes the varying weights associated with them.}
    \label{Figure 2}
\end{figure*}

% \begin{figure*}[h]
% 	\centering
% 	\includegraphics[scale=0.4]{figure/Framework.png} 
% 	\caption{The architecture of proposed \textbf{DECI} method. \( D \), \( T \), \( K \), \( L \) and \( E \) represent demographic factors, the medical note, aggregated knowledge, label, and expert respectively. In the MoE architecture, the degree of shading assigned to individual experts denotes the varying weights associated with them. }
% 	\label{Figure 3} 
% \end{figure*}

The SCM of \textbf{DECI}, which is formulated as a directed acyclic graph, is shown in \figref{Figure 2} (a). \( D \), \( T \), \( K \), \( L \), \( E_r \) and  \( E_\textit{ir} \) represent demographic factors, the medical note, aggregated knowledge, label,  relevant expert and  irrelevant expert respectively. The SCM encompasses three distinct pathways:

(1) \textbf{Aggregated Knowledge Pathway 
$Z_K$ ($(D,T) \rightarrow K \rightarrow E_r \rightarrow L
$)}: In this pathway, both D and T contribute to the formation of the knowledge node K. Subsequently, the activation of the corresponding expert within the relevant MoE framework is determined by the aggregated knowledge, culminating in the generation of the final output L; 

(2) \textbf{Direct Demographic Influence Pathway $Z_D$ ($D \rightarrow E_r \rightarrow L$)}: This route involves the prediction model relying solely on the potentially spurious correlations between the demographic inputs D and the ICD codes L, bypassing the intermediary knowledge representation.

(3) \textbf{Irrelevant Expert Activation Pathway $Z_E$ ($(D,T) \rightarrow K \rightarrow E_\textit{ir} \rightarrow L$)}: Regardless of what the fused knowledge is, each expert within the irrelevant MoE is activated uniformly, simulating the results of prior research that did not differentiate among departments. Such an undifferentiated activation may lead to the incorporation of irrelevant information into the final predictions.

With the SCM defined, we propose the framwork of \textbf{DECI} in \figref{Figure 2} (b). The desired situation for \textbf{DECI} is that the edges that bring biases are all blocked, and the prediction is based on aggregated knowledge with relevant expert. We metigate the biases by subtracting the contributions of demographic and irrelevant expert. The final score $Z_f$ for ICD codes is 
 represented as
 \begin{equation}
     Z_f =\sigma(Z_K + Z_D + Z_E)-\sigma(Z_D + Z_E)
 \end{equation}

\subsection{Aggregated Knowledge Pathway $Z_K$}
In previous studies, predictive models have been developed based solely on the input of medical notes, without taking into account potential biases introduced by demographic factors and information from other non-relevant experts. To address this limitation, a more comprehensive model is proposed, which incorporates both the medical note $T$ and demographic data $D$ to form an aggregated knowledge $K$ for making predictions. This process can be mathematically expressed as follows:

\begin{equation}
    E_K = \text{Enc}(t_T, t_D)
\end{equation}
\begin{equation}
    H_K = \text{Attention}(E_K)
\end{equation}
Here, \(t_T\) and \(t_D\) denote the encoded representations of the medical note and the demographic information, respectively. The attention mechanism is employed to generate a label-specific document representation, denoted as \(H_K\), from the token-level embeddings \(E_K\).

A considerable body of research \citep{moe1,moe2} posits that FFNs are capable of storing task-relevant knowledge. In line with this, we integrates a set of \(F_N\) FFNs, each serving as an expert in a specific department, collectively represented as \(\{FFN_i | i \in {1, 2, \ldots, F_N}\}\). These experts are coordinated through a gating mechanism, denoted as \(G()\), which acts as a decision-making entity, distributing the importance among the different experts according to the relevance of their knowledge to the aggregated representation. We implement the gating mechanism through a trainable FFN with a softmax activation function. 

In the aggregated knowledge pathway, the effect denoted as \(Z_K\) is articulated through the following formulation:
\begin{equation}
    Z_K = \sum_{i=1}^{F_N} G(H_K)_i \cdot FFN_i(H_K)
\end{equation}
where \(G(H_K)_i\) represents the output of the gating function, indicating the weight assigned to each expert based on the aggregated knowledge representation \(H_K\), effectively modeling the inter-departmental relevance.

\subsection{Direct Demographic Influence Pathway $Z_D$}
The predictive model exhibits a propensity to generate outcomes that may be unduly influenced by spurious correlations between ICD codes $L$ and demographic variables $D$, such as gender and age. To investigate the direct impact of these demographic factors on the predictions, we construct a counterfactual scenario through an intervention in which the causal link from the aggregated knowledge of the $D-T$ pairs to the final outcome is severed. Only the text about the demographic variables $D$ is input into the model to emulate the condition where knowledge about the $D-T$ relationship is absent. 

In the direct demographic influence pathway, the effect denoted as \(Z_D\) is articulated through the following formulation:
\begin{equation}
    Z_D = \sum_{i=1}^{F_N} G(H_D)_i \cdot FFN_i(H_D)
\end{equation}
where \(H_D\) denotes the document representation specific to the demographic factors $D$, derived solely from the corresponding textual inputs. This setup allows for an isolated examination of the demographic factors' influence on the model's output.

\subsection{Irrelevant Expert Activation Pathway $Z_E$}
In the porcess of ICD code assignment by human coders, the process typically leverages the specialized expertise and knowledge specific to the coders' respective departments. However, existing literature\citep{MSMN,2023Rare-ICD}, has not sufficiently explored the variability in proficiency and domain-specific competencies among these individuals. As a result, the framework being discussed assigns equal weight to the contributions from all experts. This uniform weighting approach may inadvertently introduce extraneous information from less pertinent experts, thereby potentially introducing a bias into the coding process.

Within the irrelevant expert contribution mechanism, the effect denoted as \(Z_E\) is formulated as follows:
\begin{equation}
    Z_E = \sum_{i=1}^{F_N} \frac{1}{F_N} \cdot FFN_i(H_K)
\end{equation}
wherein each expert's contribution within the framework is equally weighted with a factor of \(\frac{1}{F_N}\).

\subsection{Training and Inference}
The predictions for aggregated knowledge pathway, the direct demographic influence pathway and the irrelevant expert activation pathway are defined as:
\begin{equation}
    L_K = \sigma\left(Z_K\right)
\end{equation}
\begin{equation}
    L_D = \sigma\left(Z_D\right)
\end{equation}
\begin{equation}
    L_E = \sigma\left(Z_E\right)
\end{equation}

During the training phase, as biases are mainly involved in  demographic factors and irrelevant expert, we expect that the model captures such biases so that they can be reduced via the subtraction scheme. Motivated by this, we encourage the model to make predictions in the direct demographic influence pathway and the irrelevant expert activation pathway at training stage. And Our framework is fine-tuned by minimizing the cross-entropy loss across the different predictions. The objective function for the training process can be expressed as:
\begin{equation} 
\mathcal{L} = \mathcal{L}_K + \alpha \cdot \mathcal{L}_{D} + \beta \cdot \mathcal{L}_{E}
\end{equation}
where \(\mathcal{L}_K\), \(\mathcal{L}_D\), and \(\mathcal{L}_E\) denote the cross-entropy losses associated with \(L_K\), \(L_D\), and \(L_E\), respectively, while \(\alpha\) and \(\beta\) are parameters that regulate the relative importance of the demographic and expert pathways.

For the inference stage, we subtract the contributions of demographic and expert biases. The final prediction is operationalized through the following equation:
\begin{equation} 
L_f = \sigma(Z_f)
\end{equation}

\section{Experiments}
\subsection{Datasets}
In our study, we employ the MIMIC-III dataset \citep{Mimic3_dataset}, a comprehensive and de-identified repository that contains in excess of 40,000 patient records from the intensive care units at Beth Israel Deaconess Medical Center. This dataset is characterized by its broad spectrum of information, including but not limited to, laboratory results, vital sign measurements, medication details, and clinical documentation. Our research emphasis is on the prediction of ICD codes for discharge summaries, which are intricately associated with individual hospitalization episodes.

The discharge summaries within the MIMIC-III database are professionally coded using one or more ICD-9 codes, denoting the specific diagnoses and procedures conducted during each hospital stay. The dataset incorporates a total of 8,921 distinct ICD-9 codes, of which 6,918 pertain to diagnoses and 2,003 to procedures. It is noteworthy that the dataset accounts for patients with multiple admissions, each accompanied by separate discharge summaries. To maintain consistency with previous studies and to achieve an even distribution of patient data across training, validation, and test datasets, we have implemented the data partitioning strategy as suggested by \citep{caml}.

Two primary configurations of the MIMIC-III dataset are available: MIMIC-III Full and MIMIC-III 50. The MIMIC-III Full configuration includes all 8,921 ICD-9 codes and consists of 47,719, 1,631, and 3,372 discharge summaries designated for training, development, and testing, respectively. Conversely, the MIMIC-III 50 subset focuses on the 50 most prevalent ICD-9 codes, with 8,067, 1,574, and 1,730 discharge summaries allocated for the respective stages of model evaluation.

\subsection{Evaluation Metrics}
The evaluation framework adopted in this study is aligned with the methodological principles outlined by \citep{MSMN}. Our primary performance indicators encompass Micro and Macro AUC (area under the ROC curve), as well as Micro and Macro F1 scores. Additionally, we utilize Precision at K (Precision@K) to further assess the model's effectiveness. Specifically, for the MIMIC-III 50 dataset, we report the Precision@5 (P@5) metric, whereas for the more extensive MIMIC-III Full dataset, the Precision@8 (P@8) measure is employed.

\subsection{Baselines}
We compare our proposed method with the following baseline models :

\textbf{CAML} \citep{caml}: A pioneering model that uses a label attention layer to generate label-specific representations.

\textbf{MultiResCNN} \citep{multirescnn}: This model employs a multi-filter CNN and a residual block to capture diverse patterns in medical notes, and it also includes a label attention mechanism for code-specific representations.

\textbf{LAAT} \citep{LAAT}: LAAT introduces a new label attention mechanism over BiLSTM hidden states to learn attention scores for each medical code.

\textbf{Fusion} \citep{Fusion}: Fusion integrates multi-CNN, Transformer encoder, and label attention to improve ICD coding performance and accuracy.

\textbf{MSMN} \citep{MSMN}: MSMN utilizes code synonyms from a medical knowledge graph, along with multi-head attention and an LSTM encoder, to enhance code representation.

\textbf{Rare-ICD} \citep{2023Rare-ICD}: This model leverages a relation-enhanced code encoder to better handle rare ICD codes by modeling relationships between different codes.

\subsection{Implementation Details}
We employed the word2vec algorithm to pre-train word embeddings of dimensionality 100, utilizing a corpus comprised of medical notes and code descriptions. Concerning the training regimen for both the baseline models and their respective adaptations under the proposed DECI framework, it is important to note that the optimizer settings and other hyperparameters except $\alpha$ and $\beta$ were maintained in accordance with the default configurations recommended for the baseline models. Specifically, the values of $\alpha$ and $\beta$ are established according different models. 

For the MIMIC-III dataset in its 50 configuration, the training procedure was executed on a single NVIDIA-V100 GPU equipped with 32GB of memory. Conversely, when addressing the MIMIC-III Full dataset, the computational demands necessitated the deployment of a dual-GPU setup, featuring two NVIDIA-V100 GPUs, each providing 32GB of memory, to facilitate the training process.

\subsection{Results}
\begin{table*}
\centering
\begin{tblr}{
  width = \linewidth,
  colspec = {Q[165]Q[83]Q[73]Q[83]Q[73]Q[69]Q[83]Q[73]Q[83]Q[73]Q[69]},
  row{2} = {c},
  row{3} = {c},
  cell{1}{1} = {r=3}{},
  cell{1}{2} = {c=5}{0.381\linewidth,c},
  cell{1}{7} = {c=5}{0.381\linewidth,c},
  cell{2}{2} = {c=2}{0.156\linewidth},
  cell{2}{4} = {c=2}{0.156\linewidth},
  cell{2}{6} = {r=2}{},
  cell{2}{7} = {c=2}{0.156\linewidth},
  cell{2}{9} = {c=2}{0.156\linewidth},
  cell{2}{11} = {r=2}{},
  cell{4}{2} = {c},
  cell{4}{3} = {c},
  cell{4}{4} = {c},
  cell{4}{5} = {c},
  cell{4}{6} = {c},
  cell{4}{7} = {c},
  cell{4}{8} = {c},
  cell{4}{9} = {c},
  cell{4}{10} = {c},
  cell{4}{11} = {c},
  cell{5}{2} = {c},
  cell{5}{3} = {c},
  cell{5}{4} = {c},
  cell{5}{5} = {c},
  cell{5}{6} = {c},
  cell{5}{7} = {c},
  cell{5}{8} = {c},
  cell{5}{9} = {c},
  cell{5}{10} = {c},
  cell{5}{11} = {c},
  cell{6}{2} = {c},
  cell{6}{3} = {c},
  cell{6}{4} = {c},
  cell{6}{5} = {c},
  cell{6}{6} = {c},
  cell{6}{7} = {c},
  cell{6}{8} = {c},
  cell{6}{9} = {c},
  cell{6}{10} = {c},
  cell{6}{11} = {c},
  cell{7}{2} = {c},
  cell{7}{3} = {c},
  cell{7}{4} = {c},
  cell{7}{5} = {c},
  cell{7}{6} = {c},
  cell{7}{7} = {c},
  cell{7}{8} = {c},
  cell{7}{9} = {c},
  cell{7}{10} = {c},
  cell{7}{11} = {c},
  cell{8}{2} = {c},
  cell{8}{3} = {c},
  cell{8}{4} = {c},
  cell{8}{5} = {c},
  cell{8}{6} = {c},
  cell{8}{7} = {c},
  cell{8}{8} = {c},
  cell{8}{9} = {c},
  cell{8}{10} = {c},
  cell{8}{11} = {c},
  cell{9}{2} = {c},
  cell{9}{3} = {c},
  cell{9}{4} = {c},
  cell{9}{5} = {c},
  cell{9}{6} = {c},
  cell{9}{7} = {c},
  cell{9}{8} = {c},
  cell{9}{9} = {c},
  cell{9}{10} = {c},
  cell{9}{11} = {c},
  cell{10}{2} = {c},
  cell{10}{3} = {c},
  cell{10}{4} = {c},
  cell{10}{5} = {c},
  cell{10}{6} = {c},
  cell{10}{7} = {c},
  cell{10}{8} = {c},
  cell{10}{9} = {c},
  cell{10}{10} = {c},
  cell{10}{11} = {c},
  cell{11}{2} = {c},
  cell{11}{3} = {c},
  cell{11}{4} = {c},
  cell{11}{5} = {c},
  cell{11}{6} = {c},
  cell{11}{7} = {c},
  cell{11}{8} = {c},
  cell{11}{9} = {c},
  cell{11}{10} = {c},
  cell{11}{11} = {c},
  cell{12}{2} = {c},
  cell{12}{3} = {c},
  cell{12}{4} = {c},
  cell{12}{5} = {c},
  cell{12}{6} = {c},
  cell{12}{7} = {c},
  cell{12}{8} = {c},
  cell{12}{9} = {c},
  cell{12}{10} = {c},
  cell{12}{11} = {c},
  cell{13}{2} = {c},
  cell{13}{3} = {c},
  cell{13}{4} = {c},
  cell{13}{5} = {c},
  cell{13}{6} = {c},
  cell{13}{7} = {c},
  cell{13}{8} = {c},
  cell{13}{9} = {c},
  cell{13}{10} = {c},
  cell{13}{11} = {c},
  cell{14}{2} = {c},
  cell{14}{3} = {c},
  cell{14}{4} = {c},
  cell{14}{5} = {c},
  cell{14}{6} = {c},
  cell{14}{7} = {c},
  cell{14}{8} = {c},
  cell{14}{9} = {c},
  cell{14}{10} = {c},
  cell{14}{11} = {c},
  cell{15}{2} = {c},
  cell{15}{3} = {c},
  cell{15}{4} = {c},
  cell{15}{5} = {c},
  cell{15}{6} = {c},
  cell{15}{7} = {c},
  cell{15}{8} = {c},
  cell{15}{9} = {c},
  cell{15}{10} = {c},
  cell{15}{11} = {c},
  vline{2,7} = {1-15}{},
  hline{1,14} = {-}{0.08em},
  hline{3} = {2-5,7-10}{},
  hline{4,6,8,10,12,14,16} = {-}{},
}
Model & \textbf{MIMIC-III Full} & & & & & \textbf{MIMIC-III 50} & & & & \\
& Auc & & F1 & & P@8 & Auc &  & F1 & & P@5 \\
& Macro & Micro & Macro & Micro & & Macro & Micro & Macro & Micro & \\
CAML        
& 88.0~ & 98.3~ & 5.7~ & 50.2~ & 69.8~ 
& 87.3~ & 90.6~ & 49.6~ & 60.3~ & 60.5~ \\
+DECI       
& \textbf{90.1~} & \textbf{98.4~} & \textbf{5.6~} & \textbf{52.6~} & \textbf{70.5}~
& \textbf{87.3~} & \textbf{90.6~} & \textbf{53.6~} & \textbf{62.2~} & \textbf{61.1}~ \\

MultiResCNN 
& 90.5~ & 98.6~ & 7.6~ & 55.1~ & 73.8~ 
& 89.7~ & 92.5~ & 59.8~ & 66.8~ & 63.3~ \\
+DECI       
& \textbf{90.8~} & \textbf{98.6~} & \textbf{8.0~} & \textbf{57.1~} & \textbf{73.9}~
& \textbf{90.1~} & \textbf{92.8~} & \textbf{61.5~} & \textbf{67.3~} & \textbf{63.4}~ \\

LAAT       
& 91.9~ & 98.8~ & 9.9~ & 57.5~ & 73.8~ 
& 92.5~ & 94.6~ & 66.1~ & 71.6~ & 67.1~ \\
+DECI       
& \textbf{94.4~} & \textbf{99.1~} & \textbf{10.4~} & \textbf{57.5~} & \textbf{73.9~} 
& \textbf{92.6~} & \textbf{94.6~} & \textbf{67.3~} & \textbf{71.6~} & \textbf{67.1~}\\ 

Fusion      
& 91.5~  & 98.7~ & 8.3~ & 55.4~ & 73.6~ 
& 90.3~  & 93.1~ & 60.4~ & 67.8~ & 63.8~ \\
+DECI       
& \textbf{91.6~} & \textbf{98.7}~ & \textbf{8.3~} & \textbf{56.8~} & \textbf{74.5~}
& \textbf{90.3~} & \textbf{93.2}~ & \textbf{60.4~} & \textbf{67.8~} & \textbf{64.2~} \\

MSMN        
& 95.0~ & 99.2~ & 10.3~ & 58.2~ & 74.9~ 
& 92.7~ & 94.6~ & 67.4~ & 71.7~ & 67.4~ \\
+DECI       
& \textbf{95.0~} & \textbf{99.2~} & \textbf{10.4~} & \textbf{58.7~} & \textbf{75.0~} 
& \textbf{92.9~} & \textbf{94.7~} & \textbf{67.8~} & \textbf{71.7~} & \textbf{67.4~}\\

Rare-ICD     
& 94.7~  & 99.1~ & 10.5~ & 58.1~ & 74.5~  
& 91.9~  & 94.2~ & 64.2~ & 70.8~ & 66.5~ \\
+DECI       
& \textbf{94.8~} & \textbf{99.3}~ & \textbf{10.6~} & \textbf{58.3~} & \textbf{74.6~}
& \textbf{92.0~} & \textbf{94.2}~ & \textbf{65.6~} & \textbf{70.8~} & \textbf{66.5~} \\
\end{tblr}
\caption{Results on the MIMIC-III Full and MIMIC-III 50 datasets.}
	\label{Table_50}
\end{table*}

Table \ref{Table_50} provides a comparative analysis of the performance outcomes between the baseline models and the proposed DECI framework when applied to both the MIMIC-III Full and MIMIC-III 50 datasets.

A general trend evident from the data is that incorporating the DECI framework leads to improved performance across all baseline models. This observation underscores the utility and importance of addressing demographic and expert biases through the proposed framework. Specifically, the substantial enhancements in F1 and Precision@K scores serve as compelling evidence of the effectiveness of the DECI framework in mitigating the impact of the direct demographic influence pathway.

Furthermore, an examination of the performance metrics, particularly at the micro level, reveals valuable insights into the efficacy of the DECI framework in neutralizing the effects of the irrelevant expert activation pathway. Overall, our DECI framework effectively addresses demographic bias and expert bias in this specific context, thereby enhancing the predictive capabilities of the underlying models.

\subsection{Ablation Study}
\begin{table}
\centering
\begin{tblr}{
  width = \linewidth,
  colspec = {Q[369]Q[117]Q[106]Q[117]Q[106]Q[98]},
  cells = {c},
  cell{1}{1} = {r=2}{},
  cell{1}{2} = {c=2}{0.223\linewidth},
  cell{1}{4} = {c=2}{0.223\linewidth},
  cell{1}{6} = {r=2}{},
  hline{1-7} = {-}{},
  hline{2} = {2-5}{},
}
Model                 & Auc   &       & F1    &       & P@5   \\
                      & Macro & Micro & Macro & Micro &  \\
MSMN        
& 92.7~ & 94.6~ & 67.4~ & 71.7~ &67.4~  \\
MSMN+DECI             & 92.9~ & 94.7~ & 67.8~ & 71.7~ & 67.4~ \\
w/o $Z_D$               & 
92.7~ & 94.6~ & 67.6~ & 71.6~ & 67.2~  \\ 
w/o $Z_E$~              & 92.7~ & 94.7~ & 67.8~ & 71.7~ & 67.0~
\end{tblr}
\caption{Ablation results on the MIMIC-3 50 dataset. }
\label{Table ablation} 
\end{table}

To comprehensively assess the impact of each individual component, we carried out a series of ablation experiments on MSMN+DECI. The results of these experiments are detailed in Table \ref{Table ablation}. From these results, it can be observed that:

\textbf{Impact of the Direct Demographic Influence Pathway:} Upon exclusion of the direct demographic influence pathway (indicated as w/o $Z_D$ in Table \ref{Table ablation}), there is a notable decline in performance across all metrics, with a particularly pronounced effect on the P@5 metric. This suggests that the removal of the causal impact associated with the direct demographic influence pathway significantly diminishes the model's effectiveness.

\textbf{Impact of the Irrelevant Expert Activation Pathway:} In cases where the irrelevant expert activation pathway is not accounted for (w/o $Z_E$ in Table \ref{Table ablation}), a marked reduction in performance is observed, particularly affecting the P@5 metric. However, other performance indicators remain relatively stable. This indicates that mitigating the causal influence of the irrelevant expert activation pathway plays a crucial role in maintaining the overall performance of the model.

\subsection{Case Study}
\begin{figure}[h]
    \centering
    \includegraphics[width=0.5\textwidth]{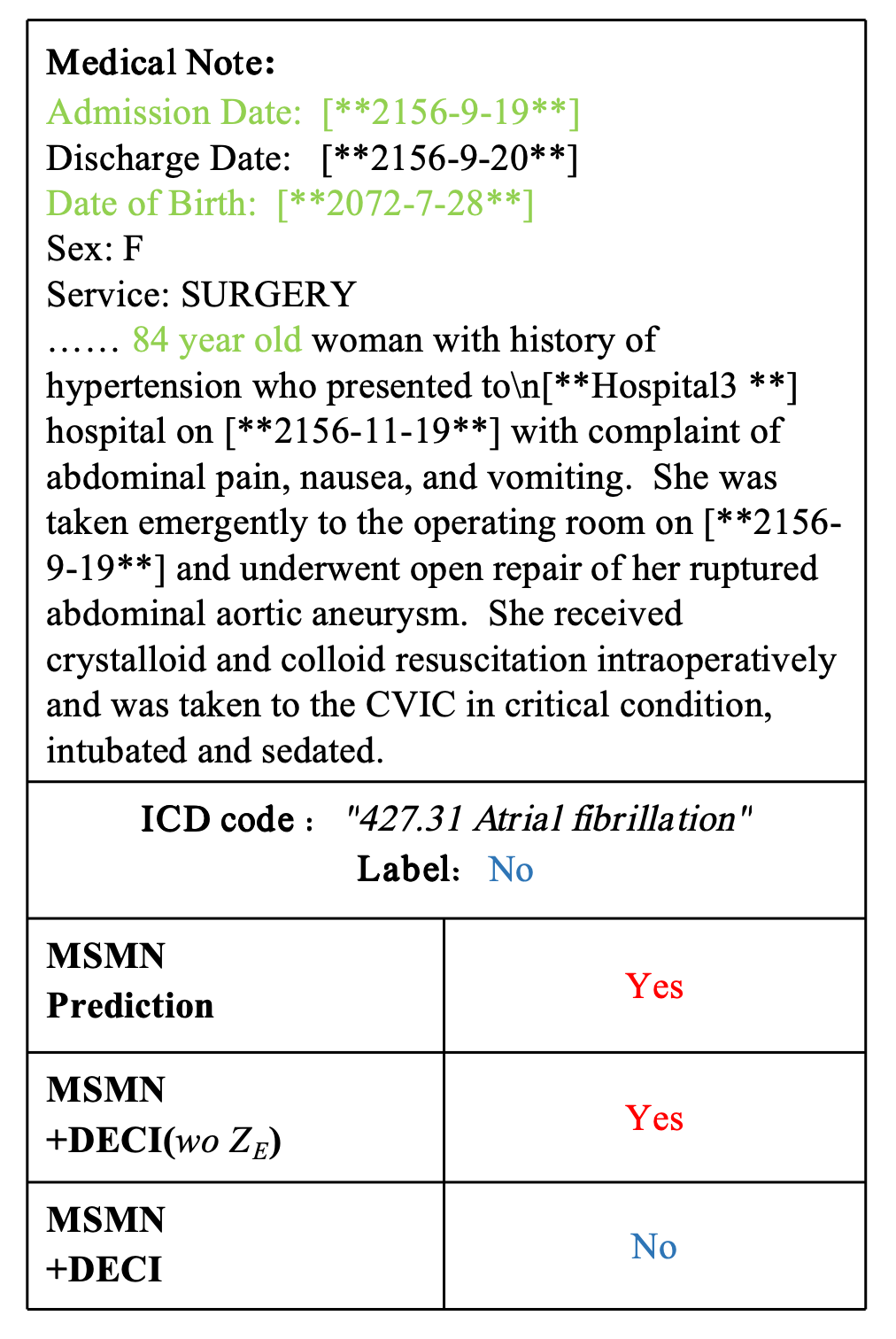}
    \caption{A representative instance where our proposed method DECI outputs correct veracity prediction while baselines MSMN make mistakes. The words highlighted in green represent terms found about advanced age in medical note.}
    \label{Figure case}
\end{figure}
In this section, we design some case studies to further analyze the advantages of our proposed method DECI on a qualitative aspect. We aim to compare the performance of the model and its counterparts within our DECI framework at an instance level. We choose the MSMN model to carry out the analysis. Specifically, we select representative examples from the dataset Hard that are correctly classified using our method while mistakenly predicted by baselines.

As illustrated in \figref{Figure case}, the primary source of bias in the given instance can be attributed to specific textual elements: "Admission Date: [**2156-9-19**]", "Date of Birth: [**2072-7-28**]", and "84 year old". These phrases strongly suggest that the patient belongs to an elderly demographic. Due to the higher co-occurrence frequency between advanced age and the ICD code "427.31 Atrial fibrillation" in the training data, the model exhibits a predisposition to overlook pertinent medical information and erroneously assign this ICD code based on the patient's age. Our methodology addresses this issue by mitigating the direct influence of demographic factors, thereby reducing such biases.

However, when the DECI approach is applied without the inclusion of expert factor $Z_E$ (MSMN+DECI(wo $Z_E$)), it results in an incorrect prediction. This outcome suggests that the irrelevant experts introduce extraneous and potentially inaccurate information into the model. Conversely, the MSMN+DECI configuration achieves accurate prediction. This success is attributed to the three pathway mechanism within our framework, which effectively captures and adjusts for both demographic and expert biases during the training phase. Subsequently, during the inference phase, these biases are systematically subtracted, ensuring a more precise and unbiased prediction.
\section{Conclusion}
In this paper, we introduce an innovative approach to mitigate \textbf{D}emographic and \textbf{E}xpert biases in ICD coding through \textbf{C}ausal \textbf{I}nference. To our knowledge, this is the pioneering effort specifically targeting the mitigation of both demographic and expert biases within this domain. Our contribution offers a unique causality-based framework for ICD coding, which elucidates the prediction process via three distinct pathways. Empirical evaluations on a standard benchmark dataset reveal that our proposed method outperforms several state-of-the-art models, thereby underscoring the efficacy and potential impact of our approach.
\section{Limitation}
In the proposed DECI framework, it is necessary to compute the causal effects associated with both the direct demographic influence pathway and the ostensibly irrelevant expert activation pathway throughout the training and inference stages. This computational requirement necessitates a somewhat greater allocation of GPU resources and an extended computational time.

Moreover, our approach mandates meticulous fine-tuning for each baseline scenario. This process involves the adjustment of hyperparameters, including but not limited to \(\alpha\) and \(\beta\), as well as the architecture of the experts, in order to optimize performance.

% Bibliography entries for the entire Anthology, followed by custom entries
%\bibliography{anthology,custom}
% Custom bibliography entries only
\bibliography{custom}

\begin{thebibliography}{32}
\providecommand{\natexlab}[1]{#1}

\bibitem[{Biswas et~al.(2021)Biswas, Pham, and Zhang}]{transicd}
Biplob Biswas, Thai-Hoang Pham, and Ping Zhang. 2021.
\newblock \href {https://doi.org/10.1007/978-3-030-77211-6_56} {Transicd: Transformer based code-wise attention model for explainable icd coding}.
\newblock In \emph{Artificial Intelligence in Medicine}, pages 469--478, Cham. Springer International Publishing.

\bibitem[{Bottle and Aylin(2008)}]{bill}
Alex Bottle and Paul Aylin. 2008.
\newblock \href {https://doi.org/10.1111/j.1475-6773.2007.00742.x} {Intelligent information: a national system for monitoring clinical performance}.
\newblock \emph{Health services research}, 43(1p1):10--31.

\bibitem[{Chen et~al.(2023)Chen, Li, Xi, Yu, and Xiong}]{2023Rare-ICD}
Jiamin Chen, Xuhong Li, Junting Xi, Lei Yu, and Haoyi Xiong. 2023.
\newblock \href {https://doi.org/10.18653/v1/2023.clinicalnlp-1.43} {Rare codes count: Mining inter-code relations for long-tail clinical text classification}.
\newblock In \emph{Clinical Natural Language Processing Workshop}.

\bibitem[{Choi et~al.(2016)Choi, Bahadori, Schuetz, Stewart, and Sun}]{monitor}
Edward Choi, Mohammad~Taha Bahadori, Andy Schuetz, Walter~F. Stewart, and Jimeng Sun. 2016.
\newblock \href {https://proceedings.mlr.press/v56/Choi16.html} {Doctor ai: Predicting clinical events via recurrent neural networks}.
\newblock In \emph{Proceedings of the 1st Machine Learning for Healthcare Conference}, volume~56 of \emph{Proceedings of Machine Learning Research}, pages 301--318, Northeastern University, Boston, MA, USA. PMLR.

\bibitem[{Falter et~al.(2024)Falter, Godderis, Scherrenberg, Kizilkilic, Xu, Mertens, Jansen, Legroux, Kindermans, Sinnaeve, Neven, and Dendale}]{transformer2-24}
Maarten Falter, Dries Godderis, Martijn Scherrenberg, Sevda~Ece Kizilkilic, Linqi Xu, Marc Mertens, Jan Jansen, Pascal Legroux, Hanne Kindermans, Peter Sinnaeve, Frank Neven, and Paul Dendale. 2024.
\newblock \href {https://doi.org/10.1093/ehjdh/ztae008} {{Using natural language processing for automated classification of disease and to identify misclassified ICD codes in cardiac disease}}.
\newblock \emph{European Heart Journal - Digital Health}, 5(3):229--234.

\bibitem[{Fan et~al.(2024)Fan, Messmer, and Jaggi}]{moe1}
Dongyang Fan, Bettina Messmer, and Martin Jaggi. 2024.
\newblock \href {https://arxiv.org/abs/2402.13089} {Towards an empirical understanding of moe design choices}.
\newblock \emph{Preprint}, arXiv:2402.13089.

\bibitem[{Gomes et~al.(2024)Gomes, Coutinho, and Martins}]{2024AccurateAW}
Gonccalo Gomes, Isabel Coutinho, and Bruno Martins. 2024.
\newblock \href {https://arxiv.org/abs/2402.03172} {Accurate and well-calibrated icd code assignment through attention over diverse label embeddings}.
\newblock In \emph{Conference of the European Chapter of the Association for Computational Linguistics}.

\bibitem[{hui Hou et~al.(2024)hui Hou, kang Wang, nan Wang, qiang Wang, and Xiao}]{rnn2024}
Wen hui Hou, Xiao kang Wang, Ya~nan Wang, Jian qiang Wang, and Fei Xiao. 2024.
\newblock \href {https://doi.org/10.1016/j.eswa.2024.123519} {Modelling long medical documents and code associations for explainable automatic icd coding}.
\newblock \emph{Expert Systems with Applications}, 249:123519.

\bibitem[{Jacobs et~al.(1991)Jacobs, Jordan, Nowlan, and Hinton}]{MOE}
Robert~A. Jacobs, Michael~I. Jordan, Steven~J. Nowlan, and Geoffrey~E. Hinton. 1991.
\newblock \href {https://doi.org/10.1162/neco.1991.3.1.79} {Adaptive mixtures of local experts}.
\newblock \emph{Neural Computation}, page 79–87.

\bibitem[{Johnson et~al.(2016)Johnson, Pollard, Shen, Lehman, Feng, Ghassemi, Moody, Szolovits, Anthony~Celi, and Mark}]{Mimic3_dataset}
Alistair~E.W. Johnson, Tom~J. Pollard, Lu~Shen, Li-wei~H. Lehman, Mengling Feng, Mohammad Ghassemi, Benjamin Moody, Peter Szolovits, Leo Anthony~Celi, and Roger~G. Mark. 2016.
\newblock \href {https://doi.org/10.1038/sdata.2016.35} {Mimic-iii, a freely accessible critical care database}.
\newblock \emph{Scientific Data}.

\bibitem[{Li and Yu(2020)}]{multirescnn}
Fei Li and Hong Yu. 2020.
\newblock \href {https://doi.org/10.1609/aaai.v34i05.6331} {Icd coding from clinical text using multi-filter residual convolutional neural network}.
\newblock \emph{Proceedings of the AAAI Conference on Artificial Intelligence}, page 8180–8187.

\bibitem[{Lin et~al.(2024)Lin, Tang, Ye, Cui, Zhu, Jin, Huang, Zhang, Pang, Ning, and Yuan}]{moe2}
Bin Lin, Zhenyu Tang, Yang Ye, Jiaxi Cui, Bin Zhu, Peng Jin, Jinfa Huang, Junwu Zhang, Yatian Pang, Munan Ning, and Li~Yuan. 2024.
\newblock \href {https://arxiv.org/abs/2401.15947} {Moe-llava: Mixture of experts for large vision-language models}.
\newblock \emph{Preprint}, arXiv:2401.15947.

\bibitem[{Liu et~al.(2023)Liu, Perez-Concha, Nguyen, Bennett, and Jorm}]{LIU2023102662}
Leibo Liu, Oscar Perez-Concha, Anthony Nguyen, Vicki Bennett, and Louisa Jorm. 2023.
\newblock \href {https://doi.org/10.1016/j.artmed.2023.102662} {Automated icd coding using extreme multi-label long text transformer-based models}.
\newblock \emph{Artificial Intelligence in Medicine}, 144:102662.

\bibitem[{Luo et~al.(2024)Luo, Wang, Wang, Chang, Wang, and Ma}]{2024CoRelationBA}
Junyu Luo, Xiaochen Wang, Jiaqi Wang, Aofei Chang, Yaqing Wang, and Fenglong Ma. 2024.
\newblock \href {https://doi.org/10.48550/arXiv.2402.15700} {Corelation: Boosting automatic icd coding through contextualized code relation learning}.
\newblock \emph{ArXiv}, abs/2402.15700.

\bibitem[{Luo et~al.(2021)Luo, Xiao, Glass, Sun, and Ma}]{Fusion}
Junyu Luo, Cao Xiao, Lucas Glass, Jimeng Sun, and Fenglong Ma. 2021.
\newblock \href {https://doi.org/10.18653/v1/2021.findings-acl.184} {Fusion: Towards automated icd coding via feature compression}.
\newblock In \emph{Findings of the Association for Computational Linguistics: ACL-IJCNLP 2021}.

\bibitem[{Medori and Fairon(2010)}]{rule}
Julia Medori and C{\'e}drick Fairon. 2010.
\newblock \href {https://aclanthology.org/W10-1113} {Machine learning and features selection for semi-automatic {ICD}-9-{CM} encoding}.
\newblock In \emph{Proceedings of the {NAACL} {HLT} 2010 Second Louhi Workshop on Text and Data Mining of Health Documents}, pages 84--89, Los Angeles, California, USA. Association for Computational Linguistics.

\bibitem[{Mou et~al.(2023)Mou, Ye, Wu, and Dai}]{rnn2023}
Chengjie Mou, Xuesong Ye, Jun Wu, and Weinan Dai. 2023.
\newblock \href {https://doi.org/10.1109/ICCCBDA56900.2023.10154772} {Automated icd coding based on neural machine translation}.
\newblock In \emph{2023 8th International Conference on Cloud Computing and Big Data Analytics (ICCCBDA)}, pages 495--500.

\bibitem[{Mullenbach et~al.(2018)Mullenbach, Wiegreffe, Duke, Sun, and Eisenstein}]{caml}
James Mullenbach, Sarah Wiegreffe, Jon Duke, Jimeng Sun, and Jacob Eisenstein. 2018.
\newblock \href {https://doi.org/10.18653/v1/n18-1100} {Explainable prediction of medical codes from clinical text}.
\newblock In \emph{Proceedings of the 2018 Conference of the North American Chapter of the Association for Computational Linguistics: Human Language Technologies, Volume 1 (Long Papers)}.

\bibitem[{Niu et~al.(2023)Niu, Wu, Li, and Li}]{NIU2023104396}
Kunying Niu, Yifan Wu, Yaohang Li, and Min Li. 2023.
\newblock \href {https://doi.org/10.1016/j.jbi.2023.104396} {Retrieve and rerank for automated icd coding via contrastive learning}.
\newblock \emph{Journal of Biomedical Informatics}, 143:104396.

\bibitem[{Ong et~al.(2023)Ong, Kedia, Harihar, Vupparaboina, Singh, Venkatesh, Vupparaboina, Bollepalli, and Chhablani}]{gpt2023}
Joshua Ong, Nikita Kedia, Sanjana Harihar, Sharat~Chandra Vupparaboina, Sumit~Randhir Singh, Ramesh Venkatesh, Kiran Vupparaboina, Sandeep~Chandra Bollepalli, and Jay Chhablani. 2023.
\newblock \href {https://jmai.amegroups.org/article/view/8198} {Applying large language model artificial intelligence for retina international classification of diseases (icd) coding}.
\newblock \emph{Journal of Medical Artificial Intelligence}, 6(0).

\bibitem[{Organization(1993)}]{icd-10}
World~Health Organization. 1993.
\newblock \emph{The ICD-10 classification of mental and behavioural disorders: diagnostic criteria for research}, volume~2.
\newblock World Health Organization.

\bibitem[{O’Malley et~al.(2005)O’Malley, Cook, Price, Wildes, Hurdle, and Ashton}]{manual}
Kimberly~J. O’Malley, Karon~F. Cook, Matt~D. Price, Kimberly~Raiford Wildes, John~F. Hurdle, and Carol~M. Ashton. 2005.
\newblock \href {https://doi.org/10.1111/j.1475-6773.2005.00444.x} {Measuring diagnoses: Icd code accuracy}.
\newblock \emph{Health Services Research}, 40(5p2):1620–1639.

\bibitem[{Perotte et~al.(2014)Perotte, Pivovarov, Natarajan, Weiskopf, Wood, and Elhadad}]{svm}
Adler Perotte, Rimma Pivovarov, Karthik Natarajan, Nicole Weiskopf, Frank Wood, and Noémie Elhadad. 2014.
\newblock \href {https://doi.org/10.1136/amiajnl-2013-002159} {Diagnosis code assignment: models and evaluation metrics}.
\newblock \emph{Journal of the American Medical Informatics Association}, page 231–237.

\bibitem[{Scheurwegs et~al.(2017)Scheurwegs, Cule, Luyckx, Luyten, and Daelemans}]{decision}
Elyne Scheurwegs, Boris Cule, Kim Luyckx, Léon Luyten, and Walter Daelemans. 2017.
\newblock \href {https://doi.org/10.1016/j.jbi.2017.09.004} {Selecting relevant features from the electronic health record for clinical code prediction.}
\newblock \emph{Journal of Biomedical Informatics}, 74:92–103.

\bibitem[{Shim et~al.(2022)Shim, Lowet, Luca, and Vanrumste}]{shim-etal-2022-exploratory}
Heereen Shim, Dietwig Lowet, Stijn Luca, and Bart Vanrumste. 2022.
\newblock \href {https://doi.org/10.18653/v1/2022.clinicalnlp-1.10} {An exploratory data analysis: the performance differences of a medical code prediction system on different demographic groups}.
\newblock In \emph{Proceedings of the 4th Clinical Natural Language Processing Workshop}, pages 93--102, Seattle, WA. Association for Computational Linguistics.

\bibitem[{Teng et~al.(2024)Teng, Zhang, Zhou, Hu, and Li}]{CNN2024}
Fei Teng, Quanmei Zhang, Xiaomin Zhou, Jie Hu, and Tianrui Li. 2024.
\newblock \href {https://doi.org/10.1016/j.eswa.2023.121861} {Few-shot icd coding with knowledge transfer and evidence representation}.
\newblock \emph{Expert Systems with Applications}, 238:121861.

\bibitem[{Tsui(2002)}]{epidemiological}
F.-C. Tsui. 2002.
\newblock \href {https://doi.org/10.1197/jamia.m1224} {Value of icd-9-coded chief complaints for detection of epidemics.}
\newblock \emph{Journal of the American Medical Informatics Association}, 9(90061):41S – 47.

\bibitem[{Varela et~al.(2024)Varela, Doktorchik, Wiebe, Southern, Knudsen, Mathur, Quan, and Eastwood}]{explore2024}
Lucia~Otero Varela, Chelsea Doktorchik, Natalie Wiebe, Danielle~A Southern, Søren Knudsen, Pallavi Mathur, Hude Quan, and Cathy~A Eastwood. 2024.
\newblock \href {https://doi.org/10.1177/18333583221106509} {International classification of diseases clinical coding training: An international survey}.
\newblock \emph{Health Information Management Journal}, 53(2):68--75.
\newblock PMID: 35838185.

\bibitem[{Vu et~al.(2020)Vu, Nguyen, and Nguyen}]{LAAT}
Thanh Vu, Dat~Quoc Nguyen, and Anthony Nguyen. 2020.
\newblock \href {https://doi.org/10.24963/ijcai.2020/461} {A label attention model for icd coding from clinical text}.
\newblock In \emph{Proceedings of the Twenty-Ninth International Joint Conference on Artificial Intelligence}.

\bibitem[{Xie et~al.(2024)Xie, Li, Yuan, Guan, Jiang, Guo, and Peng}]{2024KnowledgebasedDP}
Jing Xie, Xin Li, Ye~Yuan, Yi~Guan, Jingchi Jiang, Xitong Guo, and Xin Peng. 2024.
\newblock \href {https://doi.org/10.1016/j.knosys.2024.111395} {Knowledge-based dynamic prompt learning for multi-label disease diagnosis}.
\newblock \emph{Knowledge-Based Systems}, 286:111395.

\bibitem[{Yang et~al.(2023)Yang, Kwon, Yao, and Yu}]{yang2023multi}
Zhichao Yang, Sunjae Kwon, Zonghai Yao, and Hong Yu. 2023.
\newblock \href {https://doi.org/10.1609/aaai.v37i4.25668} {Multi-label few-shot icd coding as autoregressive generation with prompt}.
\newblock In \emph{Proceedings of the AAAI Conference on Artificial Intelligence}, volume~37, pages 5366--5374.

\bibitem[{Yuan et~al.(2022)Yuan, Tan, and Huang}]{MSMN}
Zheng Yuan, Chuanqi Tan, and Songfang Huang. 2022.
\newblock \href {https://aclanthology.org/2022.acl-short.91} {Code synonyms do matter: Multiple synonyms matching network for automatic {ICD} coding}.
\newblock In \emph{Proceedings of the 60th Annual Meeting of the Association for Computational Linguistics (Volume 2: Short Papers)}, pages 808--814, Dublin, Ireland. Association for Computational Linguistics.

\end{thebibliography}

\appendix

\end{document}